\pdfoutput=1
\documentclass[11pt]{article}

\usepackage{EMNLP2023}
\usepackage{times}
\usepackage{latexsym}
\usepackage{makecell}
\usepackage{multirow}
\usepackage{comment}

\usepackage[T1]{fontenc}
\usepackage[utf8]{inputenc}

\usepackage{microtype}

\usepackage{booktabs}
\usepackage{graphicx}
\usepackage{subcaption}
\usepackage{amsmath}
\usepackage{algorithm}
\usepackage{algorithmic}
\usepackage{verbatim}

\title{GeneGPT: Augmenting Large Language Models with Domain Tools for Improved Access to Biomedical Information}

\author{
\textbf{Qiao Jin}$^{\clubsuit}$, \textbf{Yifan Yang}$^{\clubsuit\heartsuit}$, \textbf{Qingyu Chen}$^{\clubsuit}$, \textbf{Zhiyong Lu}$^{\clubsuit}$ \\
$^{\clubsuit}$ %\text{National Center for Biotechnology Information (NCBI)}\\ 
\text{National Library of Medicine, National Institutes of Health} \\
$^{\heartsuit}$ \text{University of Maryland, College Park} \\
\texttt{\{qiao.jin, yifan.yang3, qingyu.chen, zhiyong.lu\}@nih.gov}}

\begin{document}
\maketitle

\begin{abstract}
While large language models (LLMs) have been successfully applied to various tasks, they still face challenges with hallucinations.
Augmenting LLMs with domain-specific tools such as database utilities can facilitate easier and more precise access to specialized knowledge.
In this paper, we present GeneGPT, a novel method for teaching LLMs to use the Web APIs of the National Center for Biotechnology Information (NCBI) for answering genomics questions.
Specifically, we prompt Codex to solve the GeneTuring tests with NCBI Web APIs by in-context learning and an augmented decoding algorithm that can detect and execute API calls.
Experimental results show that GeneGPT achieves state-of-the-art performance on eight tasks in the GeneTuring benchmark with an average score of 0.83, largely surpassing retrieval-augmented LLMs such as the new Bing (0.44), biomedical LLMs such as BioMedLM (0.08) and BioGPT (0.04), as well as GPT-3 (0.16) and ChatGPT (0.12).
Our further analyses suggest that: 
(1) API demonstrations have good cross-task generalizability and are more useful than documentations for in-context learning;
(2) GeneGPT can generalize to longer chains of API calls and answer multi-hop questions in GeneHop, a novel dataset introduced in this work; 
(3) Different types of errors are enriched in different tasks, providing valuable insights for future improvements.
\end{abstract}

\section{Introduction}
Large language models (LLMs) such as PaLM \cite{chowdhery2022palm} and GPT-4 \cite{gpt4} have shown great success on a wide range of general-domain Natural Language Processing (NLP) tasks.
They also achieve state-of-the-art (SOTA) performance on domain-specific tasks like biomedical question answering \cite{singhal2022large, lievin2022can, nori2023capabilities}. 
However, since there is no intrinsic mechanism for auto-regressive LLMs to ``consult'' with any source of truth, they can generate plausible-sounding but incorrect content \cite{ji2023survey}.
To tackle the hallucination issue, various studies have been proposed to augment LLMs \cite{mialon2023augmented} by either conditioning them on retrieved relevant content \cite{guu2020retrieval, lewis2020retrieval, borgeaud2022improving} or allowing them to use other external tools such as program APIs \cite{gao2022pal, parisi2022talm, schick2023toolformer, qin2023tool}.

In this work, we propose to teach LLMs to use the Web APIs of the National Center for Biotechnology Information (NCBI).
NCBI provides API access to its entire biomedical databases and tools, including Entrez Programming Utilities (E-utils) and Basic Local Alignment Search Tool (BLAST) URL API \cite{altschul1990basic,schuler1996entrez,sayers2019database}.
Enabling LLMs to use NCBI Web APIs can provide easier and more precise access to biomedical information, especially for users who are inexperienced with the database systems.
More importantly,  Web APIs can relieve users from locally implementing functionalities, maintaining large databases, and heavy computation burdens because the only requirement for using Web APIs is an internet connection. 

\begin{figure*}[h]
  \centering
  \includegraphics[width=\linewidth]{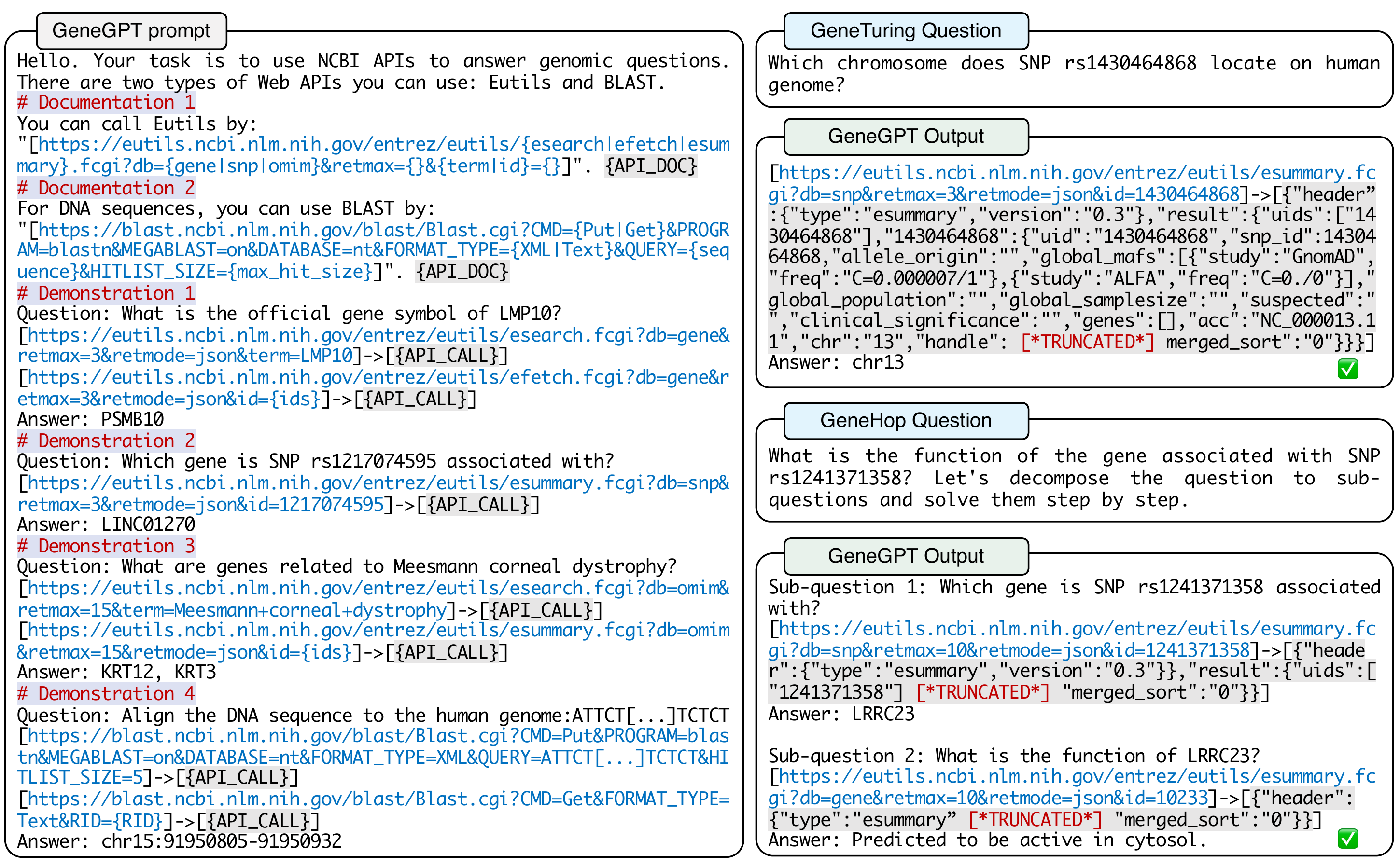}
  \caption{\textbf{Left:} GeneGPT uses NCBI Web API documentations and demonstrations in the prompt for in-context learning. \textbf{Right:} Examples of GeneGPT answering GeneTuring and GeneHop questions with NCBI Web APIs.}
  \label{fig:example}
\end{figure*}

We introduce GeneGPT, a novel method that prompts Codex \cite{chen2021evaluating} to use NCBI Web APIs by in-context learning \cite{brown2020language}.
GeneGPT consists of two main modules: (a) a specifically designed prompt that consists of documentations and demonstrations of API usage, and (b) an inference algorithm that integrates API calls in the Codex decoding process.
We evaluate GeneGPT on GeneTuring \cite{hou2023geneturing}, a question answering (QA) benchmark for genomics, and compare GeneGPT to a variety of other LLMs such as the new Bing\footnote{\url{https://www.bing.com/new}}, ChatGPT\footnote{\url{https://chat.openai.com/}}, and BioGPT \cite{luo2022biogpt}.
GeneGPT achieves the best performance on eight GeneTuring tasks with an average score of 0.83, which is remarkably higher than the previous SOTA (0.44 by New Bing).
In addition, we systematically characterize GeneGPT and find that: 
(1) API demonstrations are more useful than documentations for in-context learning;
(2) GeneGPT generalizes to longer chains of subquestion decomposition and API calls with simple demonstrations; 
(3) GeneGPT makes specific errors that are enriched for each task.

In summary, our contributions are three-fold:
\begin{enumerate}
  \item We introduce GeneGPT, a novel method that uses NCBI Web APIs to answer biomedical questions. To the best of our knowledge, this is the first study on augmenting LLMs with domain-specific Web API tools.
  \item GeneGPT achieves SOTA performance on 8 tasks in the GeneTuring benchmark, largely outperforming previous best results by 88\% (0.83 v.s. 0.44 set by the new Bing).
  \item We conduct experiments to further characterize GeneGPT, including ablation, probing, and error analyses. We also contribute a novel GeneHop dataset, and use it to show that GeneGPT can perform chain-of-thought API calls to answer multi-hop genomics questions.
\end{enumerate}

\section{GeneGPT}
In this section, we first introduce the general functions and syntax of NCBI Web APIs (\S\ref{api}).
We then describe two key components of GeneGPT: its prompt design for in-context learning (\S\ref{prompt}) and the inference algorithm (\S\ref{inference}). 

\subsection{NCBI Web APIs} \label{api}
We utilize NCBI Web APIs of E-utils\footnote{\url{https://www.ncbi.nlm.nih.gov/books/NBK25501/}} that provide access to biomedical databases and the BLAST tool\footnote{\url{https://ncbi.github.io/blast-cloud/dev/api.html}} for DNA sequence alignment.  
Web API calls are implemented by the \texttt{urllib} library in Python.

\paragraph{E-utils.} It is the API for accessing the Entrez portal \cite{schuler1996entrez}, which is a system that covers 38 NCBI databases of biomedical data such as genes and proteins \cite{sayers2019database}. 
The E-utils API provides a fixed URL syntax for rapidly retrieving such biomedical information. 
Specifically, the base URL for an E-utils request is ``\url{https://eutils.ncbi.nlm.nih.gov/entrez/eutils/{function}.fcgi}'', where \texttt{function} can be \texttt{esearch}, \texttt{efetch}, or \texttt{esummary}.
Typically, the user first calls \texttt{esearch} to get the unique database identifiers of a given query term.
Then, \texttt{efetch} or \texttt{esummary} can be called to get the full records or text summaries of a given list of identifiers returned by \texttt{esearch}. 
Important arguments in the URL request include the search term or ids (\texttt{term} or \texttt{id}), the database to use (\texttt{db}), the maximum number of returned items (\texttt{retmax}), and the return format (\texttt{retmode}).

\paragraph{BLAST URL API.} BLAST takes as input a sequence of nucleotides or amino acids and finds the most similar sequences in the database \cite{altschul1990basic, boratyn2013blast}.
The results can be used to infer relationships between sequences or identify members of gene families. 
The BLAST API allows users to submit queries to find regions of similarities between nucleotide or protein sequences to existing databases using the BLAST algorithm on NCBI servers. 
The base URL for the BLAST URL API is ``\url{https://blast.ncbi.nlm.nih.gov/blast/Blast.cgi}''. 
By sending different parameters to this API, the user can submit and retrieve queries that are executed by NCBI web servers. Every call to the API must include a \texttt{CMD} parameter that defines the type of the call. When submitting queries using \texttt{CMD=Put}, the user can specify the querying database with the \texttt{DATABASE} parameter, the searching program with the \texttt{PROGRAM} parameter, and the query sequence with the \texttt{QUERY} parameter. 
The user will get an \texttt{RID} after the \texttt{CMD=Put} API call, and can make another API call with the \texttt{Get} command and the returned \texttt{RID} to retrieve its BLAST results.

\begin{table}[!h]
\small
\centering
\begin{tabular}{llcc}
\toprule
\textbf{Comp.} & \textbf{Documentation} & \textbf{Database} & \textbf{Function} \\
\midrule
Dc.1 & E-utils & \makecell{\texttt{gene},\\\texttt{snp},\\\texttt{omim}} & \makecell{\texttt{esearch},\\\texttt{efetch},\\\texttt{esummary}} \\
\midrule
Dc.2 & BLAST & \texttt{nt} & \texttt{blastn} \\
\midrule
\textbf{Comp.} & \textbf{Demonstration} & \textbf{Database} & \textbf{Function} \\
\midrule
Dm.1 & Alias & \texttt{gene} & \makecell{\texttt{esearch->}\\\texttt{efetch}} \\
\midrule
Dm.2 & Gene SNP & \texttt{snp} & \texttt{esummary} \\
\midrule
Dm.3 & Gene disease & \texttt{omim} & \makecell{\texttt{esearch->}\\\texttt{esummary}} \\
\midrule
Dm.4 & Alignment & \texttt{nt} & \texttt{blastn} \\
\bottomrule
\end{tabular}
\caption{Summary of API usage documentations (Dc.1 and Dc.2) and demonstrations (Dm.1-4) in the GeneGPT prompt. Complete texts are shown in Appendix~\ref{appendix:prompt}.}
\label{tab:shots}
\end{table}

\subsection{In-context learning} \label{prompt}
We teach an LLM to use NCBI Web APIs through in-context learning with an engineered prompt.
Figure~\ref{fig:example} shows an example of the GeneGPT prompt, which is composed of four modules: 1. an instruction; 2. API documentations; 3. API demonstrations; 4. a test question. 
The first three parts are fixed for all tasks, while the last one is task-specific.

\begin{enumerate}
  \item \textbf{Instruction:} The prompt starts with an overall task description (``Your task is to use NCBI APIs to answer genomic questions.''). It is then followed by documentations and demonstrations of API usage summarized in Table~\ref{tab:shots}.
  \item \textbf{Documentations (Dc.)} provide natural language descriptions of the API functionality, general syntax, and argument choices. We include one for the E-utils API (Dc.1) and one for the BLAST tool (Dc.2).
  \item \textbf{Demonstrations (Dm.)} are concrete examples of using NCBI Web APIs to solve questions.
  Based on questions in the GeneTuring tasks, we manually write four demonstrations that cover four functions (\texttt{esearch}, \texttt{efetch}, \texttt{esummary}, \texttt{blastn}) and four databases (\texttt{gene}, \texttt{snp}, \texttt{omim}, \texttt{nt}) of E-utils and BLAST. 
  The API URLs and the call results are marked up by ``\texttt{[ ]}'', with a special ``\texttt{->}'' symbol inserted in between that serves as an indicator for API calls.
  \item \textbf{Test question:} The specific test question is then appended to the end of the prompt.
\end{enumerate}

While the initial GeneGPT uses all documentations and demonstrations (denoted as GeneGPT-full in Table~\ref{tab:results}), we find through analyses in \S\ref{sec:ablation} that GeneGPT can work well with only two demonstrations (denoted as GeneGPT-slim) on all tasks.

\subsection{Inference algorithm} \label{inference}

\begin{algorithm}[t]
    \caption{GeneGPT inference algorithm}
    \label{algo:ncbigpt}
\begin{algorithmic}
    \STATE {\bfseries Input: } question
    \STATE {\bfseries Model: } Codex (\texttt{code-davinci-002})
    \STATE {\bfseries Output: } answer
    \STATE prompt $\gets$ header $+$ demonstrations $+$ question
    \STATE finished $\gets$ False
    \WHILE{not finished}
        \STATE next token $\gets$ Codex(prompt)
        \STATE prompt $\gets$ prompt $+$ next token
        \IF {next token is "\texttt{->}"}
            %\STATE \textcolor{blue!50}{\texttt{\# help call Web API}}
            \STATE url $\gets$ extractLastURL(prompt)
            \STATE result $\gets$ callWebAPI(url)
            %\STATE \textcolor{blue!50}{\texttt{\# append API call result}}
            \STATE prompt $\gets$ prompt $+$ result
        \ELSIF {next token is "\texttt{\textbackslash n\textbackslash n}"}
            \STATE answer $\gets$ extractAnswer(prompt)
            \STATE finished $\gets$ True
        \ENDIF
    \ENDWHILE
\end{algorithmic}
\end{algorithm}

The GeneGPT inference algorithm is briefly shown in Algorithm~\ref{algo:ncbigpt}.
Specifically, we first append the given question to the prompt (described in  \S\ref{prompt}) and feed the concatenated text to Codex (\texttt{code-davinci-002}, \citet{chen2021evaluating}) with a temperature of 0.
We choose to use Codex for two reasons: (1) it is pre-trained with code data and shows better code understanding abilities, which is crucial in generating the URLs and interpreting the raw API results; (2) its API has the longest (8k tokens) context length among all available models so that we can fit the demonstrations in.

We discontinue the text generation process when the special ``\texttt{->}'' symbol is detected, which is the indication for an API call request.
Then we extract the last URL and call the NCBI Web API with it.
The raw execution results will be appended to the generated text, and it will be fed to Codex to continue the generation.
When ``\texttt{\textbackslash n\textbackslash n}'', an answer indicator used in the demonstrations, is generated, we will stop the inference and extract the answer after the generated ``Answer: ''.

\begin{table*}
\small
\centering
\begin{tabular}{lcccccccc}
\toprule
\multirow{2}{*}{\textbf{GeneTuring task}} & \multirow{2}{*}{\textbf{GPT-2}} & \multirow{2}{*}{\textbf{BioGPT}} & \multirow{2}{*}{\textbf{BioMedLM}} & \multirow{2}{*}{\textbf{GPT-3}} & \multirow{2}{*}{\textbf{ChatGPT}} & \multirow{2}{*}{\textbf{New Bing}} & \multicolumn{2}{c}{\textbf{GeneGPT} (ours)} \\
\cmidrule{8-9}
& & & & & & & -full & -slim \\
\midrule
\multicolumn{9}{l}{\textbf{Nomenclature}} \\
Gene alias & 0.00 & 0.00 & 0.04 & 0.09 & 0.07 & {0.66} & \underline{0.80}* & \textbf{0.84}* \\
Gene name conversion & 0.00 & 0.00 & 0.00 & 0.00 & 0.00 & \underline{0.85} & \textbf{1.00} & \textbf{1.00} \\
Average & 0.00 & 0.00 & 0.02 & 0.05 & 0.04 & {0.76} & \underline{0.90} & \textbf{0.92} \\
\midrule
\multicolumn{9}{l}{\textbf{Genomic location}} \\
Gene SNP association & 0.00 & 0.00 & 0.00 & 0.00 & 0.00 & 0.00 & \textbf{1.00}* & \textbf{1.00} \\
Gene location & 0.01 & 0.04 & 0.12 & 0.09 & 0.09 & {0.61} &  \underline{0.62} &  \textbf{0.66} \\
SNP location & 0.03 & \underline{0.05} & 0.01 & 0.02 & {0.05} & 0.01 & \textbf{1.00} & \underline{0.98} \\
Average & 0.01 & 0.03 & 0.04 & 0.04 & 0.05 & {0.21} & \underline{0.87} & \textbf{0.88} \\
\midrule
\multicolumn{9}{l}{\textbf{Functional analysis}} \\
Gene disease association & 0.00 & 0.02 & 0.16 & 0.34 & 0.31 & \textbf{0.84} & \underline{0.76}* & 0.66 \\
Protein-coding genes & 0.00 & 0.18 & 0.37 & 0.70 & 0.54 & \underline{0.97} & {0.76} & \textbf{1.00} \\
Average & 0.00 & 0.10 & 0.27 & 0.52 & 0.43 & \textbf{0.91} & {0.76} & \underline{0.84} \\
\midrule
\multicolumn{9}{l}{\textbf{Sequence alignment}} \\
DNA to human genome & 0.02 & \underline{0.07} & 0.03 & 0.00 & 0.00 & 0.00 & \textbf{0.44}* & \textbf{0.44}* \\
DNA to multiple species & 0.02 & 0.00 & 0.00 & {0.20} & 0.00 & 0.00 & \underline{0.86} & \textbf{0.88}\\
Average & 0.02 & 0.04 & 0.02 & {0.10} & 0.00 & 0.00 & \underline{0.65} & \textbf{0.66} \\
\midrule
\textbf{Overall average} & 0.00 & 0.04 & 0.08 & 0.16 & 0.12 & {0.44} & \underline{0.80} & \textbf{0.83}\\
\bottomrule
\end{tabular}
\caption{Performance of GeneGPT compared to other LLMs on the GeneTuring benchmark. *One-shot learning for GeneGPT. \textbf{Bolded} and \underline{underlined} numbers denote the highest and second-highest performance, respectively.}
\label{tab:results}
\end{table*}

\section{Experiments}
\subsection{GeneTuring} \label{geneturing}
The GeneTuring benchmark \cite{hou2023geneturing} contains 12 tasks, and each task has 50 question-answer pairs.
We use 9 GeneTuring tasks that are related to NCBI resources to evaluate the proposed GeneGPT model, and the QA samples are shown in Appendix~\ref{appendix:geneturing_samples}.
The chosen tasks are classified into four modules and briefly described in this section.

\paragraph{Nomenclature:} This is about gene names. 
We use the gene alias task and the gene name conversion task, where the objective is to find the official gene symbols for their non-official synonyms.

\paragraph{Genomics location:} The tasks are about the locations of genes, single-nucleotide polymorphism (SNP), and their relations.
We include the gene location, SNP location, and gene SNP association tasks.
The first two tasks ask for the chromosome locations (e.g., ``chr2'') of a gene or an SNP, and the last one asks for related genes for a given SNP.

\paragraph{Functional analysis:} It asks for gene functions.
We use the gene disease association task where the goal is to return related genes for a given disease, and the protein-coding genes task which asks whether a gene is a protein-coding gene or not.

\paragraph{Sequence alignment:} The tasks query specific DNA sequences.
We use the DNA sequence alignment to human genome task and the DNA sequence alignment to multiple species task.
The former maps an DNA sequence to a specific human chromosome, while the latter maps an DNA sequence to a specific species (e.g. ``zebrafish'').

\subsection{Compared methods} 
We evaluate two settings of GeneGPT, a full setting (GeneGPT-full) where all prompt components are used, as well as a slim setting (GeneGPT-slim) inspired by our ablation and probing analyses (\S\ref{sec:ablation}) where only Dm.1 and Dm.4 are used.

We compare GeneGPT with various baselines evaluated by \citet{hou2023geneturing}, 
including general-domain GPT-based \cite{radford2018improving} LLMs such as GPT-2 \cite{radford2019language}, GPT-3 (\texttt{text-davinci-003}) \cite{brown2020language}, and ChatGPT\footnote{\url{https://chat.openai.com/} (Jan 31 version).}, 
GPT-2-sized biomedical domain-specific LLMs such as BioGPT \cite{luo2022biogpt} and BioMedLM\footnote{\url{https://crfm.stanford.edu/2022/12/15/biomedlm.html}, previously known as PubMedGPT.},
as well as the new Bing\footnote{\url{https://www.bing.com/new}}, a retrieval-augmented LLM that has access to relevant web pages retrieved by Bing.

\subsection{Evaluation} \label{eval}
For the performance of the compared methods, we directly use the results reported in the original benchmark that are manually evaluated.

To evaluate our proposed GeneGPT method, we follow the general criteria but perform automatic evaluations.
Specifically, we only consider \textit{exact} matches between model predictions and the ground truth as correct predictions for all nomenclature and genomics location tasks.
For the gene disease association task, we measure the recall as in the original dataset but based on \textit{exact} individual gene matches.
For the protein-coding genes task and the DNA sequence alignment to multiple species task, we also consider \textit{exact} matches as correct after applying a simple vocabulary mapping that converts model-predicted ``yes''/``no'' to ``TRUE''/``NA'' and Latin species names to their informal names (e.g., ``\textit{Saccharomyces cerevisiae}'' to ``yeast''), respectively.
For the DNA sequence alignment to human genome task, we give correct chromosome mapping but incorrect position mapping a score of 0.5 (e.g., chr8:7081648-7081782 v.s. chr8:1207812-1207946), since the original task does not specify a reference genome.
Overall, our evaluation of GeneGPT is more strict than the original evaluation of other LLMs in \citet{hou2023geneturing}, which performs manual evaluation and might consider non-exact matches as correct.

\subsection{Main results}
Table~\ref{tab:results} shows the performance of GeneGPT on the GeneTuring tasks in comparison with other LLMs.
For GeneGPT, tasks with ``*'' in Table~\ref{tab:results} are one-shot where one instance is used as API demonstration, and the other tasks are zero-shot.
For the compared LLMs, all tasks are zero-shot.

\paragraph{Nomenclature:} GeneGPT achieves state-of-the-art (SOTA) performance on both the one-shot gene alias task with an accuracy of 0.84 and the zero-shot gene name conversion task with an accuracy of 1.00.
On average, GeneGPT outperforms New Bing by a large margin (0.92 v.s. 0.76).
All other GPT models have accuracy scores of less than 0.10 on the nomenclature tasks.

\paragraph{Genomic location:} GeneGPT also achieves SOTA performance on all genomic location tasks, including the gene SNP association task (1.00) gene location task (0.66) and the SNP location task (1.00).
While the New Bing is comparable to GeneGPT on gene location (0.61 v.s. 0.66), its performance on the two SNP-related tasks is close to 0.
Again, most other LLMs score less than 0.10.
Notably, while all genomics location tasks are zero-shot for GeneGPT-slim, it performs comparably to GeneGPT-full which uses one gene SNP association demonstration.
This indicates that API demonstrations have strong cross-task generalizability.

\begin{figure*}[!ht]
  \centering
  \begin{subfigure}{0.48\textwidth}
    \includegraphics[width=\linewidth]{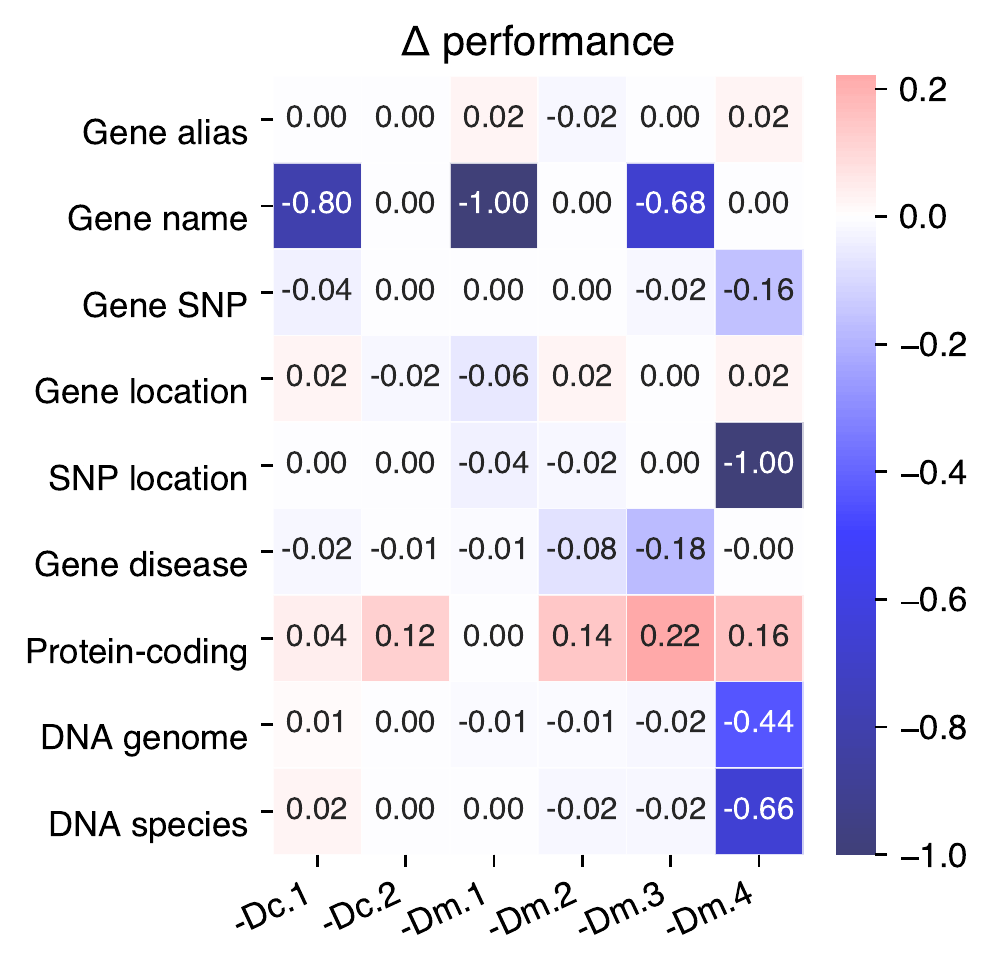}
  \end{subfigure}
  \hfill
  \begin{subfigure}{0.48\textwidth}
    \includegraphics[width=\linewidth]{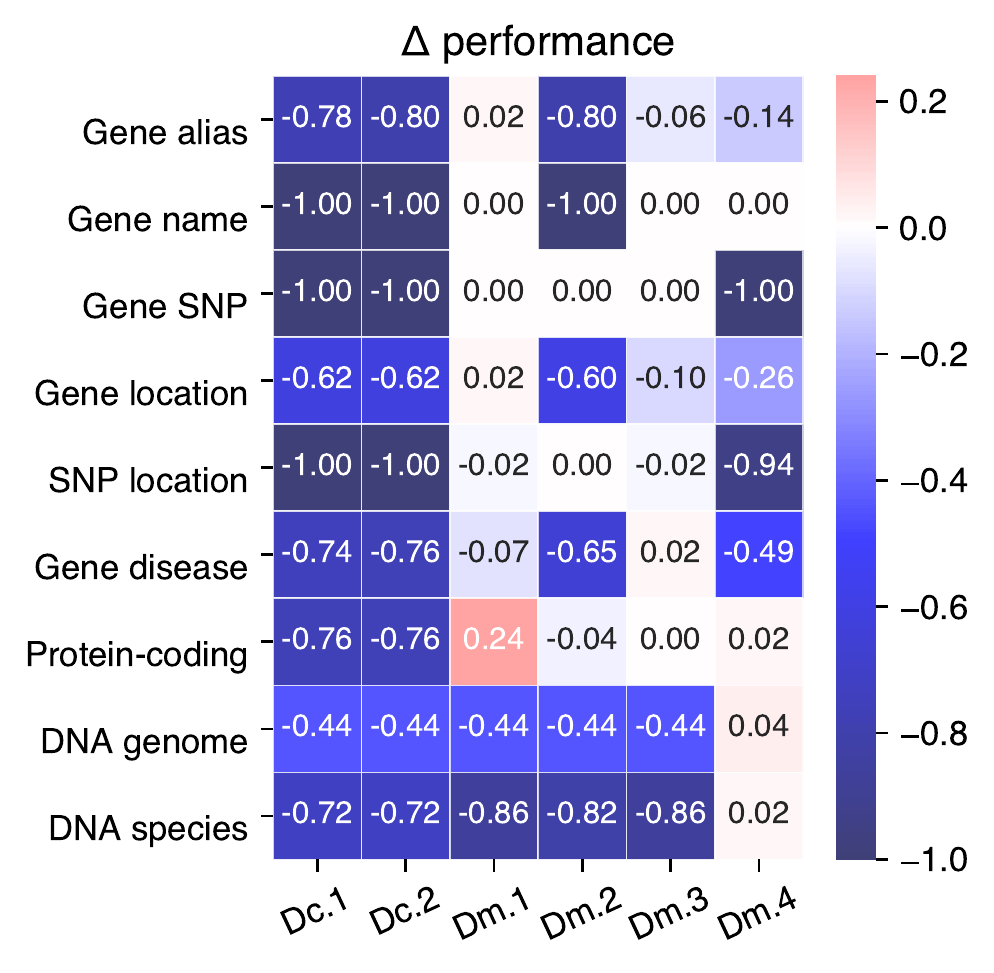}
  \end{subfigure}
  \caption{Performance changes of the ablation (\textbf{left}) and probing (\textbf{right}) experiments as compared to GeneGPT-full.}
  \label{fig:ablation}
\end{figure*}

\paragraph{Functional analysis:}
The new Bing performs better functional analysis tasks than the proposed GeneGPT (average score: 0.91 v.s. 0.84), which is probably because many web pages related to gene functions can be retrieved by the Bing search engine.
We also note that other LLMs, especially GPT-3 and ChatGPT, perform moderately well and much better than they perform on other tasks.
This might also be due to the fact that many gene-function-related texts are included in their pre-training corpora.

\paragraph{Sequence alignment:}
GeneGPT performs much better with an average score of 0.66 than all other models including the new Bing (0.00), which essentially fails on the sequence alignment tasks.
This is not very surprising since sequence alignment is easy with the BLAST tool, but almost impossible for an auto-regressive LLM even with retrieval augmentation as the input sequences are too specific to be indexed by a search engine.

Although evaluated under a more strict setting (\S\ref{eval}), GeneGPT achieves a macro-average performance of 0.83 which is much higher than other compared LLMs including New Bing (0.44).
Overall, GeneGPT achieves new SOTA performance on all 2 one-shot tasks and 6 out of 7 zero-shot tasks and is outperformed by New Bing only on the gene disease association task.

\section{Discussions}
We have shown that GeneGPT largely surpasses various LLMs on the GeneTuring benchmark.
In this section, we further characterize GeneGPT by studying three research questions (RQ):

\textbf{RQ1}: What is the importance of each prompt component in GeneGPT?

\textbf{RQ2}: Can GeneGPT answer multi-hop questions by chain-of-thought API calls?

\textbf{RQ3}: What types of errors does GeneGPT make on each studied task?

\subsection{RQ1: Component importance}
\label{sec:ablation}
We conduct ablation and probing experiments to study the importance of individual prompt components, including 2 documentations (Dc.1, Dc.2) and 4 demonstrations (Dm.1-4) described in \S\ref{prompt}.

For ablation tests, we remove each component from GeneGPT-full and then evaluate the prompt.
The results are shown in Figure~\ref{fig:ablation} (left).
Notably, the performance on the DNA to genome and species alignment tasks is only significantly decreased without the BLAST demonstration (Dm.4), but not affected by the ablation of the BLAST documentation (Dc.2).
While the ablations of other components decrease the performance, most only affect one relevant task (e.g., Dm.1 and gene name conversion), which indicates a high level of redundancy of the prompt components.

For the probing experiments, we evaluate GeneGPT with only one prompt component to study the individual capability.
The results are shown in Figure~\ref{fig:ablation} (right).
Overall, GeneGPT with only one documentation (Dc.1 or Dc.2) fails on all tasks.
Surprisingly, with only one demonstration of the gene alias task (Dm.1) in the prompt, GeneGPT is able to perform comparably to GeneGPT-full on all tasks except the alignment ones.
On the other hand, GeneGPT with only the BLAST demonstration (Dm.4) performs well on the two alignment tasks, which is somehow expected.
These results suggest that GeneGPT with only two demonstrations (Dm.1 and Dm.4) in the prompt can generalize to all tasks in the GeneTuring benchmark.
We denote this as GeneGPT-slim, and results in Table~\ref{tab:results} show that with only two demonstrations, it outperforms the GeneGPT-full and achieves state-of-the-art overall results on GeneTuring.

\subsection{RQ2: Multi-hop QA on GeneHop}
Questions in the GeneTuring benchmark are single-hop and just require one step of reasoning, e.g., ``Which gene is SNP rs983419152 associated with?''.
However, many real-world biomedical questions are multi-hop that need more steps to answer \cite{jin2022biomedical}.
For example, to answer ``What is the function of the gene associated with SNP rs983419152?'', the model should first get the associated gene name and then find its functions.

\begin{figure}[!ht]
    \framebox{
    \parbox{0.45\textwidth}{
    \small
    \colorbox{blue!20}{\textbf{{GeneHop question (SNP gene function)}:}} \newline
    What is the function of the gene associated with SNP rs1241371358? Let's decompose the question to sub-questions and solve them step by step. \newline
    %\newline
    \colorbox{blue!20}{\textbf{GeneGPT output:}} \newline
    Sub-question 1: Which gene is SNP rs1241371358 associated with? \newline
    [\url{https://eutils.ncbi.nlm.nih.gov/entrez/eutils/esummary.fcgi?db=snp&retmax=10&retmode=json&id=1241371358}]\\->[\colorbox{black!20}{API call results}] \newline
    Answer: LRRC23 \newline
    %\newline
    Sub-question 2: What is the function of LRRC23? \newline
    [\url{https://eutils.ncbi.nlm.nih.gov/entrez/eutils/esummary.fcgi?db=gene&retmax=10&retmode=json&id=10233}]\\->[\colorbox{black!20}{API call results}] \newline
    \colorbox{green!20}{\textbf{Answer: Predicted to be active in cytosol. (correct)}}
    }
    }
    \caption{GeneGPT uses chain-of-thought API calls to answer a multi-hop question in GeneHop.} 
    \label{fig:multi_1}
\end{figure}

To test GeneGPT's capability of answering multi-hop questions, we present \textbf{GeneHop}, a novel dataset that contains three new multi-hop QA tasks based on the GeneTuring benchmark: (a) \textbf{SNP gene function}, which asks for the function of the gene associated with a given SNP.  (b) \textbf{Disease gene location}, where the task is to list the chromosome locations of the genes associated with a given disease.
(c) \textbf{Sequence gene alias}, which asks for the aliases of the gene that contains a specific DNA sequence.
Each task in GeneHop contains 50 questions, and the collection pipeline is detailed in Appendix~\ref{appendix:genehop_collection}.
%For tasks (a) and (b), we use GeneGPT without BLAST prompt components (Dc.2 and Dm.4).
%For task (c), we use GeneGPT without Dm.2.
For all tasks, we append the chain-of-thought instruction ``Let's decompose the question to sub-questions and solve them step by step.'' after the test question \cite{wei2022chain}.

Figure~\ref{fig:multi_1} shows an example of GeneGPT to answer Task (a). 
In this case, GeneGPT successfully decomposes the multi-hop question into two sub-questions, and the sub-question 2 is based on the answer of the sub-question 1.
Interestingly, GeneGPT uses a shortcut to answer sub-question 2: instead of first calling \texttt{esearch} and then calling \texttt{esummary}, GeneGPT finds the gene id in the API call results of sub-question 1 and directly calls \texttt{esummary}.
This capability is not shown in the prompt but elicited by chain-of-thought API calls.

\begin{figure}[!ht]
    \framebox{
    \parbox{0.45\textwidth}{
    \small
    \colorbox{blue!20}{\textbf{{GeneHop question (Disease gene location)}:}} \newline
    List chromosome locations of the genes related to Cleft palate with ankyloglossia. Let's decompose the question to sub-questions and solve them step by step. \newline 
    %\newline
    \colorbox{blue!20}{\textbf{GeneGPT output:}} \newline
    Sub-question 1: What is the OMIM id of Cleft palate with ankyloglossia? \newline
    [\url{https://eutils.ncbi.nlm.nih.gov/entrez/eutils/esearch.fcgi?db=omim&retmax=20&retmode=json&sort=relevance&term=Cleft+palate+with+ankyloglossia}]->[\colorbox{black!20}{API call results}] \newline
    Answer: 303400 \newline
    %\newline
    Sub-question 2: What are genes related to Cleft palate with ankyloglossia? \newline
    [\url{https://eutils.ncbi.nlm.nih.gov/entrez/eutils/esummary.fcgi?db=omim&retmax=20&retmode=json&id=303400}]\\->[\colorbox{black!20}{API call results}] \newline
    Answer: CPX \newline 
    %\newline
    Sub-question 3: What is the chromosome location of CPX? \newline 
    [\url{https://eutils.ncbi.nlm.nih.gov/entrez/eutils/esearch.fcgi?db=gene&retmax=20&retmode=json&sort=relevance&term=CPX}]->[\colorbox{black!20}{API call results}] \newline
    [\url{https://eutils.ncbi.nlm.nih.gov/entrez/eutils/esummary.fcgi?db=gene&retmax=20&retmode=json&id=64877,1377,7758276,100818277,100166185,10948718,10849362,9700326,3180310,3180309,1192676,1371,50945,10682,12892,1037,4336951,100216262,380161,108700304}]\\->[\colorbox{black!20}{API call results}] \newline
    \colorbox{green!20}{\textbf{Answer: Xq21.1 (correct)}}
    }
    }
    \caption{GeneGPT uses chain-of-thought API calls to answer a multi-hop question in GeneHop.} 
    \label{fig:multi_2}
\end{figure}

Figure~\ref{fig:multi_2} shows another example of GeneGPT answering Task (b), where GeneGPT successfully decomposes the multi-hop question and correctly calls the required APIs.
Notably, the answering chain involves 3 sub-questions and 4 API calls, which are longer than all in-context demonstrations (1 single-hop question and 2 API calls at most).
This ability to generalize to longer chains of thought is an important aspect of GeneGPT's flexibility and usefulness for real-world applications.

\begin{table}[!ht]
\small
\centering
\begin{tabular}{lcc}
\toprule
\textbf{GeneHop Task} & \textbf{New Bing} & \textbf{GeneGPT} \\
\midrule
SNP gene function & 0.00 & \textbf{0.55} \\
Disease gene location & \textbf{0.71} & 0.67 \\ 
Sequence gene alias & 0.00 & \textbf{0.28} \\
\midrule
Average & 0.24 & \textbf{0.50} \\
\bottomrule
\end{tabular}
\caption{Performance of multi-hop QA on GeneHop. We only compare GeneGPT with New Bing since other LLMs cannot even answer single-hop questions well.}
\label{tab:multihop}
\end{table}

We manually evaluate the results predicted by GeneGPT and compare it to the new Bing, which is the only baseline LLM that performs well on the single-hop GeneTuring benchmark due to its retrieval augmentation feature.
The evaluation criteria are described in Appendix~\ref{appendix:genehop_evaluation}.
As shown in Table~\ref{tab:multihop}, while the new Bing outperforms GeneGPT on the disease gene location task, it is mostly using webpages that contain both the disease and location information without multi-hop reasoning.
The new Bing fails to perform the other 2 tasks since the input information (SNP or sequence) is not indexed by Bing and can only be found in specialized databases.
GeneGPT, on the other hand, performs moderately well on all 3 tasks, and achieves a much higher average score (0.50 v.s. 0.24).

\subsection{RQ3: Error analysis}
We manually study all errors made by GeneGPT and classify them into five types.
Table~\ref{tab:error} shows the count of each error type on the evaluate tasks:
\textbf{E1:} using the wrong API or not using APIs, e.g., using the \texttt{gene} instead of the \texttt{omin} database for diseases; \textbf{E2:} using the right API but wrong arguments, e.g., passing terms to \texttt{id}; \textbf{E3:} not extracting the answer in the API result, most commonly seen in gene function extraction; \textbf{E4:} right API call but results do not contain the answer, where the question is not answerable with NCBI databases; and \textbf{O} includes other unclassified errors.
Specific error examples are shown in Appendix~\ref{appendix:error}.

\begin{table}[!hb]
\small
\centering
\begin{tabular}{lccccc}
\toprule
\textbf{GeneTuring Task} & \textbf{E1} & \textbf{E2} & \textbf{E3} & \textbf{E4} & \textbf{O}\\
\midrule
Gene alias & 0 & 0 & 2 & \textcolor{red!80}{\textbf{6}} & 0 \\
Gene location & 0 & 0 & 0 & \textcolor{red!80}{\textbf{17}} & 0\\
SNP location & 0 & \textcolor{red!80}{\textbf{1}} & 0 & 0 & 0 \\
Gene disease association & \textcolor{red!80}{\textbf{15}} & 0 & 0 & 3 & 2 \\
DNA to human genome & 0 & 0 & 7 & 0 & \textcolor{red!80}{\textbf{42}} \\
DNA to multiple species & 0 & 0 & 1 & 1 & \textcolor{red!80}{\textbf{4}} \\
\midrule
\textbf{GeneHop Task} & \textbf{E1} & \textbf{E2} & \textbf{E3} & \textbf{E4} & \textbf{O}\\
\midrule
SNP gene function & 0 & 0 & \textcolor{red!80}{\textbf{29}} & 0 & 0\\
Disease gene location & 4 & \textcolor{red!80}{\textbf{7}} & 1 & 5 & 1\\ 
Sequence gene alias & 0 & \textcolor{red!80}{\textbf{30}} & 8 & 0 & 0\\
\bottomrule
\end{tabular}
\caption{Counts of GeneGPT errors on different tasks. \textbf{E1}: wrong API; \textbf{E2}: wrong arguments; \textbf{E3}: wrong comprehension; \textbf{E4}: unanswerable with API; \textbf{O}: others.}
\label{tab:error}
\end{table}

Our results suggest that different tasks have specific and enriched error types: simple tasks (alias and location) fail mostly because of \textbf{E4}; \textbf{E1} only happens in disease-related tasks; alignment tasks face more issues with BLAST interfaces and reference genomes (\textbf{O}); multi-hop tasks in GeneHop tend to have \textbf{E2} and \textbf{E3} in the reasoning chains.

\section{Related work}
\paragraph{Large language models:}
Recent studies have shown that scaling pre-trained LMs leads to performance improvement and potentially emergent abilities on various NLP tasks \cite{brown2020language, kaplan2020scaling, wei2022emergent, chowdhery2022palm, gpt4}.
However, such auto-regressive LLMs are still susceptible to hallucinations and generate erroneous content \cite{ji2023survey}.
Augmenting LLMs with external tools is a possible solution to this issue \cite{mialon2023augmented}.

\paragraph{Tool augmentation:}
Potential tools include: (1) search engines \cite{guu2020retrieval, lewis2020retrieval, borgeaud2022improving}, also known as retrieval augmentation, exemplified by New Bing; (2) program APIs by in-context learning \cite{gao2022pal, schick2023toolformer} or fine-tuning \cite{parisi2022talm, schick2023toolformer}.
We present the first study on the in-context learning abilities of documentations and demonstrations of NCBI Web APIs.

\paragraph{Biomedical question answering:}
It is an essential step in clinical decision support \cite{ely2005answering} and biomedical knowledge acquisition \cite{jin2022biomedical}.
LLMs have been successfully applied to various biomedical QA tasks that are \textit{knowledge}- or \textit{reasoning}-intensive \cite{singhal2022large, lievin2022can, nori2023capabilities}.
However, auto-regressive LLMs fail to perform \textit{data}-intensive tasks which require the model to precisely store and recite database entries, such as the GeneTuring benchmark \cite{hou2023geneturing}.
Retrieval augmentation also falls short since specialized databases are usually not indexed by commercial search engines.
GeneGPT solves this task by tool augmentation.

\section{Conclusions}
We present GeneGPT, a novel method that teaches LLMs to use NCBI Web APIs.
It achieves SOTA performance on 8 GeneTuring tasks and can perform chain-of-thought API calls.
Our results indicate that database utility tools might be superior to relevant web pages for augmenting LLMs to faithfully serve various biomedical information needs.

\bibliography{anthology, custom}

\begin{thebibliography}{29}
\expandafter\ifx\csname natexlab\endcsname\relax\def\natexlab#1{#1}\fi

\bibitem[{Altschul et~al.(1990)Altschul, Gish, Miller, Myers, and
  Lipman}]{altschul1990basic}
Stephen~F Altschul, Warren Gish, Webb Miller, Eugene~W Myers, and David~J
  Lipman. 1990.
\newblock Basic local alignment search tool.
\newblock \emph{Journal of molecular biology}, 215(3):403--410.

\bibitem[{Boratyn et~al.(2013)Boratyn, Camacho, Cooper, Coulouris, Fong, Ma,
  Madden, Matten, McGinnis, Merezhuk et~al.}]{boratyn2013blast}
Grzegorz~M Boratyn, Christiam Camacho, Peter~S Cooper, George Coulouris, Amelia
  Fong, Ning Ma, Thomas~L Madden, Wayne~T Matten, Scott~D McGinnis, Yuri
  Merezhuk, et~al. 2013.
\newblock Blast: a more efficient report with usability improvements.
\newblock \emph{Nucleic acids research}, 41(W1):W29--W33.

\bibitem[{Borgeaud et~al.(2022)Borgeaud, Mensch, Hoffmann, Cai, Rutherford,
  Millican, Van Den~Driessche, Lespiau, Damoc, Clark
  et~al.}]{borgeaud2022improving}
Sebastian Borgeaud, Arthur Mensch, Jordan Hoffmann, Trevor Cai, Eliza
  Rutherford, Katie Millican, George~Bm Van Den~Driessche, Jean-Baptiste
  Lespiau, Bogdan Damoc, Aidan Clark, et~al. 2022.
\newblock Improving language models by retrieving from trillions of tokens.
\newblock In \emph{International conference on machine learning}, pages
  2206--2240. PMLR.

\bibitem[{Brown et~al.(2020)Brown, Mann, Ryder, Subbiah, Kaplan, Dhariwal,
  Neelakantan, Shyam, Sastry, Askell et~al.}]{brown2020language}
Tom Brown, Benjamin Mann, Nick Ryder, Melanie Subbiah, Jared~D Kaplan, Prafulla
  Dhariwal, Arvind Neelakantan, Pranav Shyam, Girish Sastry, Amanda Askell,
  et~al. 2020.
\newblock Language models are few-shot learners.
\newblock \emph{Advances in neural information processing systems},
  33:1877--1901.

\bibitem[{Chen et~al.(2021)Chen, Tworek, Jun, Yuan, Pinto, Kaplan, Edwards,
  Burda, Joseph, Brockman et~al.}]{chen2021evaluating}
Mark Chen, Jerry Tworek, Heewoo Jun, Qiming Yuan, Henrique Ponde de~Oliveira
  Pinto, Jared Kaplan, Harri Edwards, Yuri Burda, Nicholas Joseph, Greg
  Brockman, et~al. 2021.
\newblock Evaluating large language models trained on code.
\newblock \emph{arXiv preprint arXiv:2107.03374}.

\bibitem[{Chowdhery et~al.(2022)Chowdhery, Narang, Devlin, Bosma, Mishra,
  Roberts, Barham, Chung, Sutton, Gehrmann et~al.}]{chowdhery2022palm}
Aakanksha Chowdhery, Sharan Narang, Jacob Devlin, Maarten Bosma, Gaurav Mishra,
  Adam Roberts, Paul Barham, Hyung~Won Chung, Charles Sutton, Sebastian
  Gehrmann, et~al. 2022.
\newblock Palm: Scaling language modeling with pathways.
\newblock \emph{arXiv preprint arXiv:2204.02311}.

\bibitem[{Ely et~al.(2005)Ely, Osheroff, Chambliss, Ebell, and
  Rosenbaum}]{ely2005answering}
John~W Ely, Jerome~A Osheroff, M~Lee Chambliss, Mark~H Ebell, and Marcy~E
  Rosenbaum. 2005.
\newblock Answering physicians' clinical questions: obstacles and potential
  solutions.
\newblock \emph{Journal of the American Medical Informatics Association},
  12(2):217--224.

\bibitem[{Gao et~al.(2022)Gao, Madaan, Zhou, Alon, Liu, Yang, Callan, and
  Neubig}]{gao2022pal}
Luyu Gao, Aman Madaan, Shuyan Zhou, Uri Alon, Pengfei Liu, Yiming Yang, Jamie
  Callan, and Graham Neubig. 2022.
\newblock Pal: Program-aided language models.
\newblock \emph{arXiv preprint arXiv:2211.10435}.

\bibitem[{Guu et~al.(2020)Guu, Lee, Tung, Pasupat, and
  Chang}]{guu2020retrieval}
Kelvin Guu, Kenton Lee, Zora Tung, Panupong Pasupat, and Mingwei Chang. 2020.
\newblock Retrieval augmented language model pre-training.
\newblock In \emph{International conference on machine learning}, pages
  3929--3938. PMLR.

\bibitem[{Hou and Ji(2023)}]{hou2023geneturing}
Wenpin Hou and Zhicheng Ji. 2023.
\newblock Geneturing tests gpt models in genomics.
\newblock \emph{bioRxiv}, pages 2023--03.

\bibitem[{Ji et~al.(2023)Ji, Lee, Frieske, Yu, Su, Xu, Ishii, Bang, Madotto,
  and Fung}]{ji2023survey}
Ziwei Ji, Nayeon Lee, Rita Frieske, Tiezheng Yu, Dan Su, Yan Xu, Etsuko Ishii,
  Ye~Jin Bang, Andrea Madotto, and Pascale Fung. 2023.
\newblock Survey of hallucination in natural language generation.
\newblock \emph{ACM Computing Surveys}, 55(12):1--38.

\bibitem[{Jin et~al.(2022)Jin, Yuan, Xiong, Yu, Ying, Tan, Chen, Huang, Liu,
  and Yu}]{jin2022biomedical}
Qiao Jin, Zheng Yuan, Guangzhi Xiong, Qianlan Yu, Huaiyuan Ying, Chuanqi Tan,
  Mosha Chen, Songfang Huang, Xiaozhong Liu, and Sheng Yu. 2022.
\newblock Biomedical question answering: a survey of approaches and challenges.
\newblock \emph{ACM Computing Surveys (CSUR)}, 55(2):1--36.

\bibitem[{Kaplan et~al.(2020)Kaplan, McCandlish, Henighan, Brown, Chess, Child,
  Gray, Radford, Wu, and Amodei}]{kaplan2020scaling}
Jared Kaplan, Sam McCandlish, Tom Henighan, Tom~B Brown, Benjamin Chess, Rewon
  Child, Scott Gray, Alec Radford, Jeffrey Wu, and Dario Amodei. 2020.
\newblock Scaling laws for neural language models.
\newblock \emph{arXiv preprint arXiv:2001.08361}.

\bibitem[{Lewis et~al.(2020)Lewis, Perez, Piktus, Petroni, Karpukhin, Goyal,
  K{\"u}ttler, Lewis, Yih, Rockt{\"a}schel et~al.}]{lewis2020retrieval}
Patrick Lewis, Ethan Perez, Aleksandra Piktus, Fabio Petroni, Vladimir
  Karpukhin, Naman Goyal, Heinrich K{\"u}ttler, Mike Lewis, Wen-tau Yih, Tim
  Rockt{\"a}schel, et~al. 2020.
\newblock Retrieval-augmented generation for knowledge-intensive nlp tasks.
\newblock \emph{Advances in Neural Information Processing Systems},
  33:9459--9474.

\bibitem[{Li{\'e}vin et~al.(2022)Li{\'e}vin, Hother, and
  Winther}]{lievin2022can}
Valentin Li{\'e}vin, Christoffer~Egeberg Hother, and Ole Winther. 2022.
\newblock Can large language models reason about medical questions?
\newblock \emph{arXiv preprint arXiv:2207.08143}.

\bibitem[{Luo et~al.(2022)Luo, Sun, Xia, Qin, Zhang, Poon, and
  Liu}]{luo2022biogpt}
Renqian Luo, Liai Sun, Yingce Xia, Tao Qin, Sheng Zhang, Hoifung Poon, and
  Tie-Yan Liu. 2022.
\newblock Biogpt: generative pre-trained transformer for biomedical text
  generation and mining.
\newblock \emph{Briefings in Bioinformatics}, 23(6).

\bibitem[{Mialon et~al.(2023)Mialon, Dess{\`\i}, Lomeli, Nalmpantis, Pasunuru,
  Raileanu, Rozi{\`e}re, Schick, Dwivedi-Yu, Celikyilmaz
  et~al.}]{mialon2023augmented}
Gr{\'e}goire Mialon, Roberto Dess{\`\i}, Maria Lomeli, Christoforos Nalmpantis,
  Ram Pasunuru, Roberta Raileanu, Baptiste Rozi{\`e}re, Timo Schick, Jane
  Dwivedi-Yu, Asli Celikyilmaz, et~al. 2023.
\newblock Augmented language models: a survey.
\newblock \emph{arXiv preprint arXiv:2302.07842}.

\bibitem[{Nori et~al.(2023)Nori, King, McKinney, Carignan, and
  Horvitz}]{nori2023capabilities}
Harsha Nori, Nicholas King, Scott~Mayer McKinney, Dean Carignan, and Eric
  Horvitz. 2023.
\newblock Capabilities of gpt-4 on medical challenge problems.
\newblock \emph{arXiv preprint arXiv:2303.13375}.

\bibitem[{OpenAI(2023)}]{gpt4}
OpenAI. 2023.
\newblock \href {https://doi.org/10.48550/arXiv.2303.08774} {{GPT-4} technical
  report}.
\newblock \emph{CoRR}, abs/2303.08774.

\bibitem[{Parisi et~al.(2022)Parisi, Zhao, and Fiedel}]{parisi2022talm}
Aaron Parisi, Yao Zhao, and Noah Fiedel. 2022.
\newblock Talm: Tool augmented language models.
\newblock \emph{arXiv preprint arXiv:2205.12255}.

\bibitem[{Qin et~al.(2023)Qin, Hu, Lin, Chen, Ding, Cui, Zeng, Huang, Xiao, Han
  et~al.}]{qin2023tool}
Yujia Qin, Shengding Hu, Yankai Lin, Weize Chen, Ning Ding, Ganqu Cui, Zheni
  Zeng, Yufei Huang, Chaojun Xiao, Chi Han, et~al. 2023.
\newblock Tool learning with foundation models.
\newblock \emph{arXiv preprint arXiv:2304.08354}.

\bibitem[{Radford et~al.(2018)Radford, Narasimhan, Salimans, Sutskever
  et~al.}]{radford2018improving}
Alec Radford, Karthik Narasimhan, Tim Salimans, Ilya Sutskever, et~al. 2018.
\newblock Improving language understanding by generative pre-training.

\bibitem[{Radford et~al.(2019)Radford, Wu, Child, Luan, Amodei, Sutskever
  et~al.}]{radford2019language}
Alec Radford, Jeffrey Wu, Rewon Child, David Luan, Dario Amodei, Ilya
  Sutskever, et~al. 2019.
\newblock Language models are unsupervised multitask learners.
\newblock \emph{OpenAI blog}, 1(8):9.

\bibitem[{Sayers et~al.(2019)Sayers, Agarwala, Bolton, Brister, Canese, Clark,
  Connor, Fiorini, Funk, Hefferon et~al.}]{sayers2019database}
Eric~W Sayers, Richa Agarwala, Evan~E Bolton, J~Rodney Brister, Kathi Canese,
  Karen Clark, Ryan Connor, Nicolas Fiorini, Kathryn Funk, Timothy Hefferon,
  et~al. 2019.
\newblock Database resources of the national center for biotechnology
  information.
\newblock \emph{Nucleic acids research}, 47(Database issue):D23.

\bibitem[{Schick et~al.(2023)Schick, Dwivedi-Yu, Dess{\`\i}, Raileanu, Lomeli,
  Zettlemoyer, Cancedda, and Scialom}]{schick2023toolformer}
Timo Schick, Jane Dwivedi-Yu, Roberto Dess{\`\i}, Roberta Raileanu, Maria
  Lomeli, Luke Zettlemoyer, Nicola Cancedda, and Thomas Scialom. 2023.
\newblock Toolformer: Language models can teach themselves to use tools.
\newblock \emph{arXiv preprint arXiv:2302.04761}.

\bibitem[{Schuler et~al.(1996)Schuler, Epstein, Ohkawa, and
  Kans}]{schuler1996entrez}
GD~Schuler, JA~Epstein, H~Ohkawa, and JA~Kans. 1996.
\newblock Entrez: molecular biology database and retrieval system.
\newblock \emph{Methods in enzymology}, 266:141--162.

\bibitem[{Singhal et~al.(2022)Singhal, Azizi, Tu, Mahdavi, Wei, Chung, Scales,
  Tanwani, Cole-Lewis, Pfohl et~al.}]{singhal2022large}
Karan Singhal, Shekoofeh Azizi, Tao Tu, S~Sara Mahdavi, Jason Wei, Hyung~Won
  Chung, Nathan Scales, Ajay Tanwani, Heather Cole-Lewis, Stephen Pfohl, et~al.
  2022.
\newblock Large language models encode clinical knowledge.
\newblock \emph{arXiv preprint arXiv:2212.13138}.

\bibitem[{Wei et~al.(2022{\natexlab{a}})Wei, Tay, Bommasani, Raffel, Zoph,
  Borgeaud, Yogatama, Bosma, Zhou, Metzler et~al.}]{wei2022emergent}
Jason Wei, Yi~Tay, Rishi Bommasani, Colin Raffel, Barret Zoph, Sebastian
  Borgeaud, Dani Yogatama, Maarten Bosma, Denny Zhou, Donald Metzler, et~al.
  2022{\natexlab{a}}.
\newblock Emergent abilities of large language models.
\newblock \emph{arXiv preprint arXiv:2206.07682}.

\bibitem[{Wei et~al.(2022{\natexlab{b}})Wei, Wang, Schuurmans, Bosma, Chi, Le,
  and Zhou}]{wei2022chain}
Jason Wei, Xuezhi Wang, Dale Schuurmans, Maarten Bosma, Ed~Chi, Quoc Le, and
  Denny Zhou. 2022{\natexlab{b}}.
\newblock Chain of thought prompting elicits reasoning in large language
  models.
\newblock \emph{arXiv preprint arXiv:2201.11903}.

\end{thebibliography}
\bibliographystyle{acl_natbib}

\newpage
\appendix
\section{GeneGPT prompt}
\label{appendix:prompt}
Here we show the exact texts of each prompt component described in Table~\ref{tab:shots}, including two documentations Dc.1 (E-utils, Figure~\ref{fig:doc_1}) and Dc.2 (BLAST, Figure~\ref{fig:doc_2}) as well as four demonstrations Dm.1 (gene alias, Figure~\ref{fig:dem_1}), Dm.2 (gene SNP association, Figure~\ref{fig:dem_2}), Dm.3 (gene disease association, Figure~\ref{fig:dem_3}), and Dm.4 (DNA to human genome alignment, Figure~\ref{fig:dem_4}).

\begin{figure*}[!hp]
    \framebox{
    \parbox{0.96\textwidth}{
    \small
    \colorbox{blue!20}{\textbf{Documentation 1 (Dc. 1)}} \newline
    You can call Eutils by: "[\url{https://eutils.ncbi.nlm.nih.gov/entrez/eutils/{esearch|efetch|esummary}.fcgi?db={gene|snp|omim}&retmax={}&{term|id}={term|id}}]".\newline
    esearch: input is a search term and output is database id(s).\newline
    efectch/esummary: input is database id(s) and output is full records or summaries that contain name, chromosome location, and other information.\newline
    Normally, you need to first call esearch to get the database id(s) of the search term, and then call efectch/esummary to get the information with the database id(s).\newline
    Database: gene is for genes, snp is for SNPs, and omim is for genetic diseases.
    }
    }
    \caption{Documentation 1 (Dc.1) of the GeneGPT prompt. Dc.1 describes the functionality, general syntax, and argument choices of the NCBI E-utils API.} 
    \label{fig:doc_1}
\end{figure*}

\begin{figure*}[!hp]
    \framebox{
    \parbox{0.96\textwidth}{
    \small
    \colorbox{blue!20}{\textbf{Documentation 2 (Dc. 2)}} \newline
    For DNA sequences, you can use BLAST by: "[\url{https://blast.ncbi.nlm.nih.gov/blast/Blast.cgi?CMD={Put|Get}&PROGRAM=blastn&MEGABLAST=on&DATABASE=nt&FORMAT_TYPE={XML|Text}&QUERY={sequence}&HITLIST_SIZE={max_hit_size}}]".\newline
    BLAST maps a specific DNA {sequence} to its chromosome location among different species.\newline
    You need to first PUT the BLAST request and then GET the results using the RID returned by PUT.
    }
    }
    \caption{Documentation 2 (Dc.2) of the GeneGPT prompt. Dc.2 describes the functionality, general syntax, and argument choices of the BLAST API.} 
    \label{fig:doc_2}
\end{figure*}

\begin{figure*}[!hp]
    \framebox{
    \parbox{0.96\textwidth}{
    \small
    \colorbox{blue!20}{\textbf{Demonstration 1 (Dm. 1)}} \newline
    Question: What is the official gene symbol of LMP10?\newline
    [\url{https://eutils.ncbi.nlm.nih.gov/entrez/eutils/esearch.fcgi?db=gene&retmax=5&retmode=json&sort=relevance&term=LMP10}]->[\textcolor{purple}{\texttt{\{"header":\{"type":"esearch","version":"0.3"\}\\,"esearchresult":\{"count":"3","retmax":"3","retstart":"0","idlist":["5699",\\"8138","19171"],"translationset":[],"translationstack":[\{"term":"LMP10[All Fields]","field":"All Fields","count":"3","explode":"N"\},"GROUP"],\\"querytranslation":"LMP10[All Fields]"\}\}}}]\newline
    [\url{https://eutils.ncbi.nlm.nih.gov/entrez/eutils/efetch.fcgi?db=gene&retmax=5&retmode=json&id=19171,5699,8138}]->[\textcolor{purple}{\texttt{1. Psmb10
    Official Symbol: Psmb10 and Name: proteasome (prosome, macropain) subunit, beta type 10 [Mus musculus (house mouse)]
    Other Aliases: Mecl-1, Mecl1
    Other Designations: proteasome subunit beta type-10; low molecular mass protein 10; macropain subunit MECl-1; multicatalytic endopeptidase complex subunit MECl-1; prosome Mecl1; proteasome (prosomome, macropain) subunit, beta type 10; proteasome MECl-1; proteasome subunit MECL1; proteasome subunit beta-2i
    Chromosome: 8; Location: 8 53.06 cM
    Annotation: Chromosome 8 NC\_000074.7 (106662360..106665024, complement)
    ID: 19171
    2. PSMB10
    Official Symbol: PSMB10 and Name: proteasome 20S subunit beta 10 [Homo sapiens (human)]
    Other Aliases: LMP10, MECL1, PRAAS5, beta2i
    Other Designations: proteasome subunit beta type-10; low molecular mass protein 10; macropain subunit MECl-1; multicatalytic endopeptidase complex subunit MECl-1; proteasome (prosome, macropain) subunit, beta type, 10; proteasome MECl-1; proteasome catalytic subunit 2i; proteasome subunit MECL1; proteasome subunit beta 10; proteasome subunit beta 7i; proteasome subunit beta-2i; proteasome subunit beta2i
    Chromosome: 16; Location: 16q22.1
    Annotation: Chromosome 16 NC\_000016.10 (67934506..67936850, complement)
    MIM: 176847
    ID: 5699
    3. MECL1
    Proteosome subunit MECL1 [Homo sapiens (human)]
    Other Aliases: LMP10, PSMB10
    This record was replaced with GeneID: 5699
    ID: 8138}}]\newline
    Answer: PSMB10
    }
    }
    \caption{Demonstration 1 (Dm.1) of the GeneGPT prompt. The instance is chosen from the gene alias task in the GeneTuring benchmark. \url{Links} are actually called Web API URLs. Readers can directly click the link and get the \textcolor{purple}{\texttt{API call result}}, which is inserted in the prompt.}
    \label{fig:dem_1}
\end{figure*}

\begin{figure*}[!hp]
    \framebox{
    \parbox{0.96\textwidth}{
    \small
    \colorbox{blue!20}{\textbf{Demonstration 2 (Dm. 2)}} \newline
    Question: Which gene is SNP rs1217074595 associated with?\newline
    [\url{https://eutils.ncbi.nlm.nih.gov/entrez/eutils/esummary.fcgi?db=snp&retmax=10&retmode=json&id=1217074595}]->[\textcolor{purple}{\texttt{\{"header":\{"type":"esummary","version":"0.3"\},\\"result":\{"uids":["1217074595"],"1217074595":\{"uid":"1217074595","snp\_id":\\1217074595,"allele\_origin":"","global\_mafs":[\{"study":"GnomAD","freq":"A=\\0.000007/1"\},\{"study":"TOPMED","freq":"A=0.000004/1"\},\{"study":"ALFA","freq":"A=\\0./0"\}],"global\_population":"","global\_samplesize":"","suspected":"","clinical\\\_significance":"","genes":[\{"name":"LINC01270","gene\_id":"284751"\}],"acc":"NC\\\_000020.11","chr":"20","handle":"GNOMAD\\,TOPMED","spdi":"NC\_000020.11:50298394:G:A","fxn\_class":"non\_coding\_transcript\\\_variant","validated":"by-frequency,by-alfa,by-cluster","docsum":"HGVS=NC\_000020\\.11:g.50298395G>A,NC\_000020.10:g.48914932G>A,NR\_034124.1:n.351G>A,NM\_001025463.1\\:c.*4G>A|SEQ=[G/A]|LEN=1|GENE=LINC01270:284751","tax\_id":9606,"orig\_build"\\:155,"upd\_build":156,"createdate":"2017/11/09 09:55","updatedate":"2022/10/13 17:11","ss":"4354715686,5091242333","allele":"R","snp\_class":"snv","chrpos":"20:\\50298395","chrpos\_prev\_assm":"20:48914932","text":"","snp\_id\_sort":"1217074595",\\"clinical\_sort":"0","cited\_sort":"","chrpos\_sort":"0050298395","merged\_sort":"0"\\\}\}\}}}]\newline
    Answer: LINC01270
    }
    }
    \caption{Demonstration 2 (Dm.2) of the GeneGPT prompt. The instance is chosen from the gene SNP association task in the GeneTuring benchmark. \url{Links} are actually called Web API URLs. Readers can directly click the link and get the \textcolor{purple}{\texttt{API call result}}, which is inserted in the prompt.} 
    \label{fig:dem_2}
\end{figure*}

\begin{figure*}[!hp]
    \framebox{
    \parbox{0.96\textwidth}{
    \small
    \colorbox{blue!20}{\textbf{Demonstration 3 (Dm. 3)}} \newline
    Question: What are genes related to Meesmann corneal dystrophy?\newline
    [\url{https://eutils.ncbi.nlm.nih.gov/entrez/eutils/esearch.fcgi?db=omim&retmax=20&retmode=json&sort=relevance&term=Meesmann+corneal+dystrophy}]->[\textcolor{purple}{\texttt{\{"header":\{"type":"esearch","version":"0.3"\},"esearchresult":\{"count":"5","ret\\max":"5","retstart":"0","idlist":["122100","618767","300778","601687","148043"],\\"translationset":[],"translationstack":[\{"term":"Meesmann[All Fields]","field":"All Fields","count":"5","explode":"N"\},\{"term":"corneal[All Fields]","field":"All Fields","count":"561","explode":"N"\},"AND",\{"term":\\"dystrophy[All Fields]","field":"All Fields","count":"1326","explode":"N"\},"AND"\\,"GROUP"],"querytranslation":"Meesmann[All Fields] AND corneal[All Fields] AND dystrophy[All Fields]"\}\}}}]\newline
    [\url{https://eutils.ncbi.nlm.nih.gov/entrez/eutils/esummary.fcgi?db=omim&retmax=20&retmode=json&id=618767,601687,300778,148043,122100}]->[\textcolor{purple}{\texttt{\{"header":\{"type":"esummary","version":"0.3"\},"result":\{"uids":["618767",\\"601687","300778","148043","122100"],"618767":\{"uid":"618767","oid":"\#618767",\\"title":"CORNEAL DYSTROPHY, MEESMANN, 2; MECD2","alttitles":"","locus":"12q13.13"\\\},"601687":\{"uid":"601687","oid":"*601687","title":"KERATIN 12, TYPE I; KRT12","alttitles":"","locus":"17q21.2"\},"300778":\{"uid":"300778","oid":"\%300778"\\,"title":"CORNEAL DYSTROPHY, LISCH EPITHELIAL; LECD","alttitles":"","locus":\\"Xp22.3"\},"148043":\{"uid":"148043","oid":"*148043","title":"KERATIN 3, TYPE II; KRT3","alttitles":"","locus":"12q13.13"\},"122100":\{"uid":"122100","oid":"\#122100"\\,"title":"CORNEAL DYSTROPHY, MEESMANN, 1; MECD1","alttitles":"","locus":"17q21.2"\\\}\}\}}}]\newline
    Answer: KRT12, KRT3
    }
    }
    \caption{Demonstration 3 (Dm.3) of the GeneGPT prompt. The instance is chosen from the gene disease association task in the GeneTuring benchmark. \url{Links} are actually called Web API URLs. Readers can directly click the link and get the \textcolor{purple}{\texttt{API call result}}, which is inserted in the prompt.}
    \label{fig:dem_3}
\end{figure*}

\begin{figure*}[!hp]
    \framebox{
    \parbox{0.97\textwidth}{
    \small
    \colorbox{blue!20}{\textbf{Demonstration 4 (Dm. 4)}} \newline
    Question: Align the DNA sequence to the human genome:ATTCTGCCTTTAGTAATTTGATGACAGAGACTTCTTGGGAA\newline 
    CCACAGCCAGGGAGCCACCCTTTACTCCACCAACAGGTGGCTTATATCCAATCTGAGAAAGAAAGAAAAAAAAA\newline
    AAAGTATTTCTCT \newline
    [\url{https://blast.ncbi.nlm.nih.gov/blast/Blast.cgi?CMD=Put&PROGRAM=blastn&MEGABLAST=on&DATABASE=nt&FORMAT_TYPE=XML&QUERY=ATTCT...TCTCT&HITLIST_SIZE=5}]->[\textcolor{purple}{\texttt{5S8YKEBH016}}] \newline
    [\url{https://blast.ncbi.nlm.nih.gov/blast/Blast.cgi?CMD=Get&FORMAT_TYPE=Text&RID=5S8YKEBH016}]->[\textcolor{purple}{\texttt{BLASTN 2.14.0+ [...]
    RID: 5S8YKEBH016
    Database: Nucleotide collection (nt)
               93,066,592 sequences; 1,138,553,367,010 total letters
    Query= 
    Length=128
                                                                       Score     E     Max
    Sequences producing significant alignments:                       (Bits)  Value  Ident
    CP034493.1 Eukaryotic synthetic construct chromosome 15            237     3e-58  100\%      
    NG\_132175.1 Homo sapiens H3K27ac-H3K4me1 hESC enhancer GRCh37\_...  237     3e-58  100\%      
    CP068263.2 Homo sapiens isolate CHM13 chromosome 15                237     3e-58  100\%      
    AP023475.1 Homo sapiens DNA, chromosome 15, nearly complete ge...  237     3e-58  100\%      
    FJ515841.1 Homo sapiens isolate SLC3A1-VI-T solute carrier org...  237     3e-58  100\%      
    ALIGNMENTS
    >CP034493.1 Eukaryotic synthetic construct chromosome 15
     CP034518.1 Eukaryotic synthetic construct chromosome 15
    Length=82521392
     Score = 237 bits (128),  Expect = 3e-58
     Identities = 128/128 (100\%), Gaps = 0/128 (0\%)
     Strand=Plus/Plus
    Query  1         ATTCTGCCTTTAGTAATTTGATGACAGAGACTTCTTGGGAACCACAGCCAGGGAGCCACC  60
                     ||||||||||||||||||||||||||||||||||||||||||||||||||||||||||||
    Sbjct  72494035  ATTCTGCCTTTAGTAATTTGATGACAGAGACTTCTTGGGAACCACAGCCAGGGAGCCACC  72494094
    Query  61        CTTTACTCCACCAACAGGTGGCTTATATCCAATCTGAGAAAGAAAGaaaaaaaaaaaaGT  120
                     ||||||||||||||||||||||||||||||||||||||||||||||||||||||||||||
    Sbjct  72494095  CTTTACTCCACCAACAGGTGGCTTATATCCAATCTGAGAAAGAAAGAAAAAAAAAAAAGT  72494154
    Query  121       ATTTCTCT  128
                     ||||||||
    Sbjct  72494155  ATTTCTCT  72494162
    >NG\_132175.1 Homo sapiens H3K27ac-H3K4me1 hESC enhancer GRCh37\_chr15:92493309-92494181 
    (LOC127830695) on chromosome 15
    Length=1073
     Score = 237 bits (128),  Expect = 3e-58
     Identities = 128/128 (100\%), Gaps = 0/128 (0\%)
     Strand=Plus/Plus
    Query  1    ATTCTGCCTTTAGTAATTTGATGACAGAGACTTCTTGGGAACCACAGCCAGGGAGCCACC  60
                ||||||||||||||||||||||||||||||||||||||||||||||||||||||||||||
    Sbjct  827  ATTCTGCCTTTAGTAATTTGATGACAGAGACTTCTTGGGAACCACAGCCAGGGAGCCACC  886
    Query  61   CTTTACTCCACCAACAGGTGGCTTATATCCAATCTGAGAAAGAAAGaaaaaaaaaaaaGT  120
                ||||||||||||||||||||||||||||||||||||||||||||||||||||||||||||
    Sbjct  887  CTTTACTCCACCAACAGGTGGCTTATATCCAATCTGAGAAAGAAAGAAAAAAAAAAAAGT  946
    Query  121  ATTTCTCT  128
                ||||||||
    Sbjct  947  ATTTCTCT  954
    >CP068263.2 Homo sapiens isolate CHM13 chromosome 15
    Length=99753195
     Score = 237 bits (128),  Expect = 3e-58
     Identities = 128/128 (100\%), Gaps = 0/128 (0\%)
     Strand=Plus/Plus
    Query  1         ATTCTGCCTTTAGTAATTTGATGACAGAGACTTCTTGGGAACCACAGCCAGGGAGCCACC  60
                     ||||||||||||||||||||||||||||||||||||||||||||||||||||||||||||
    Sbjct  89712558  ATTCTGCCTTTAGTAATTTGATGACAGAGACTTCTTGGGAACCACAGCCAGGGAGCCACC  89712617
    Query  61        CTTTACTCCACCAACAGGTGGCTTATATCCAATCTGAGAAAGAAAGaaaaaaaaaaaaGT  120
                     ||||||||||||||||||||||||||||||||||||||||||||||||||||||||||||
    Sbjct  89712618  CTTTACTCCACCAACAGGTGGCTTATATCCAATCTGAGAAAGAAAGAAAAAAAAAAAAGT  89712677
    Query  121       ATTTCTCT  128
                     ||||||||
    Sbjct  89712678  ATTTCTCT  89712685
    >AP023475.1 Homo sapiens DNA, chromosome 15, nearly complete genome
    Length=95537968
     Score = 237 bits (128),  Expect = 3e-58
     Identities = 128/128 (100\%), Gaps = 0/128 (0\%)
     Strand=Plus/Plus
    Query  1         ATTCTGCCTTTAGTAATTTGATGACAGAGACTTCTTGGGAACCACAGCCAGGGAGCCACC  60
                     ||||||||||||||||||||||||||||||||||||||||||||||||||||||||||||
    Sbjct  85572367  ATTCTGCCTTTAGTAATTTGATGACAGAGACTTCTTGGGAACCACAGCCAGGGAGCCACC  85572426
    Query  61        CTTTACTCCACCAACAGGTGGCTTATATCCAATCTGAGAAAGAAAGaaaaaaaaaaaaGT  120
                     ||||||||||||||||||||||||||||||||||||||||||||||||||||||||||||
    Sbjct  85572427  CTTTACTCCACCAACAGGTGGCTTATATCCAATCTGAGAAAGAAAGAAAAAAAAAAAAGT  85572486
    Query  121       ATTTCTCT  128
                     ||||||||
    Sbjct  85572487  ATTTCTCT  85572494
    >FJ515841.1 Homo sapiens isolate SLC3A1-VI-T solute carrier organic anion 
    transporter family member 3A1 (SLCO3A1) gene, complete cds
    Length=315834
     Score = 237 bits (128),  Expect = 3e-58
     Identities = 128/128 (100\%), Gaps = 0/128 (0\%)
     Strand=Plus/Plus
    [...]}}]\newline
    Answer: chr15:91950805-91950932
    }
    }
    \caption{Demonstration 4 (Dm.4) of the GeneGPT prompt. The instance is chosen from the DNA to human genome alignment task in the GeneTuring benchmark. \url{Links} are actually called Web API URLs. Readers can directly click the link and get the \textcolor{purple}{\texttt{API call result}}, which is inserted in the prompt. Since the BLAST Web API is about to be depricated and the first step returns an HTML page, we use a regular expression to extract the \texttt{PID} and append it to the GeneGPT input.}
    \label{fig:dem_4}
\end{figure*}

\section{GeneTuring samples}
\label{appendix:geneturing_samples}
Table~\ref{tab:geneturing_samples} shows some sample question-answer pairs from the GeneTuring benchmark.
The dataset is publicly available at \url{https://www.biorxiv.org/content/10.1101/2023.03.11.532238v1.supplementary-material}.
We use the same questions sent to the new Bing as the test questions.

\begin{table*}[hp!]
\small
\centering
\begin{tabular}{p{3.5cm}p{7.5cm}p{3.5cm}}
\toprule
\textbf{GeneTuring task} & \textbf{Question} & \textbf{Answer}\\
\midrule
\multicolumn{3}{l}{\textbf{\colorbox{black!10}{Nomenclature}}} \\
\midrule
Gene alias & What is the official gene symbol of SNAT6? & SLC38A6\\
\midrule
Gene name conversion & Convert ENSG00000215251 to official gene symbol. & FASTKD5\\
\midrule
\multicolumn{3}{l}{\colorbox{black!10}{\textbf{Genomic location}}} \\
\midrule
Gene SNP association & Which gene is SNP rs996319727 associated with? & USP39 \\
\midrule
Gene location & Which chromosome is FOXL2NB gene located on human genome? & chr3\\
\midrule
SNP location & Which chromosome does SNP rs427884 locate on human genome? & chr11 \\
\midrule
\multicolumn{3}{l}{\colorbox{black!10}{\textbf{Functional analysis}}} \\
\midrule
Gene disease association & What are genes related to Bile acid malabsorption? & SLC10A2, SLC51B \\
\midrule
Protein-coding genes & Is ATP5F1EP2 a protein-coding gene? & NA \\
\midrule
\multicolumn{3}{l}{\colorbox{black!10}{\textbf{Sequence alignment}}} \\
\midrule
DNA to human genome & Align the DNA sequence to the human genome: AGGCCC TCACCT GGAAAT TACTTA CTCATG CTTCAT GACCCA GTTCAA ATTTTG TCACCT CTGTGA AACCTT CCCTGG GCCCCG TTGATC TCCTTG AAGGCA & chr7:71368450-71368551\\
\midrule
DNA to multiple species & Which organism does the DNA sequence come from: AGGGGC AGCAAA CACCGG GACACA CCCATT CGTGCA CTAATC AGAAAC TTTTTT TTCTCA AATAAT TCAAAC AATCAA AATTGG TTTTTT CGAGCA AGGTGG GAAATT TTTCGAT & worm\\
\bottomrule
\end{tabular}
\caption{Sample question-answer pairs of the GeneTuring tasks \cite{hou2023geneturing}.}
\label{tab:geneturing_samples}
\end{table*}

\section{GeneHop collection}
\label{appendix:genehop_collection}
The GeneHop dataset contains three multi-hop tasks: SNP gene function, disease gene location, and sequence gene alias. We describe the collection of these tasks in this section.
Table~\ref{tab:genehop_samples} shows several question-answer samples from the GeneHop dataset.

\begin{table*}[hp!]
\small
\centering
\begin{tabular}{p{3.5cm}p{7.5cm}p{3.5cm}}
\toprule
\textbf{GeneHop task} & \textbf{Question} & \textbf{Answer}\\
\midrule
SNP gene function & What is the function of the gene associated with SNP rs1318850293? Let's decompose the question to sub-questions and solve them step by step. & Predicted to enable guanyl-nucleotide exchange facto
r activity. Predicted to be involved in Rho protein signal transduction. \\
\midrule
Disease gene location & List chromosome locations of the genes related to Hemolytic anemia due to phosphofructokinase deficiency. Let's decompose the question to sub-questions and solve them step by step. & 21q22.3 \\
\midrule
Sequence gene alias & What are the aliases of the gene that contains this sequence: GTAGAT GGAACT GGTAGT CAGCTG GAGAGC AGCATG GAGGCG TCCTGG GGGAGC TTCAAC GCTGAG CGGGGC TGGTAT GTCTCT GTCCAG CAGCCT GAAGAA GCGGAG GCCGA. Let's decompose the question to sub-questions and solve them step by step. & SLC38A6, NAT-1, SNAT6 \\
\bottomrule
\end{tabular}
\caption{Sample question-answer pairs of the GeneHop tasks (introduced in this work).}
\label{tab:genehop_samples}
\end{table*}

\paragraph{SNP gene function:} The question template for this task is \textit{``What is the function of the gene associated with SNP \texttt{\{snp\}}? Let's decompose the question to sub-questions and solve them step by step.''}.
We re-use the 50 \texttt{\{snp\}} from the gene SNP association task in the original GeneTuring benchmark.
The ground-truth answer of the gene function is manually annotated: 
For each SNP, we first get its corresponding gene from the annotations of the gene SNP association task.
We then check the gene information page\footnote{\url{https://www.ncbi.nlm.nih.gov/gene/}} and select its functional summary as the ground-truth answer.

\paragraph{Disease gene location:} The question template for this task is \textit{``List chromosome locations of the genes related to \texttt{\{disease\}}. Let's decompose the question to sub-questions and solve them step by step.''}.
Similarly, we re-use the 50 \texttt{\{disease\}} from the gene disease association task in the original GeneTuring benchmark.
The ground-truth list of the chromosome locations is manually annotated:
For each disease, we first get its corresponding genes from the annotations of the gene disease association task.
We then check their NCBI gene information pages and label the cytogenetics locations (e.g., 21q22.3).

\paragraph{Sequence gene alias:} The question template for this task is \textit{``What are the aliases of the gene that contains this sequence: \texttt{\{sequence\}}. Let's decompose the question to sub-questions and solve them step by step.''}.
We find the information pages of the 50 genes used in the gene alias task in the original GeneTuring benchmark.
We manually crop part of the sequence with a similar length to the sequences in the GeneTuring alignment tasks to serve as the \texttt{\{sequence\}}, and use the union of its official name and the alias set as the ground-truth answer.

\section{GeneHop evaluation}
\label{appendix:genehop_evaluation}

\paragraph{SNP gene function:} 
We manually evaluate all answers predicted by GeneGPT and the New Bing against the ground-truth gene functions.
New Bing answers with \textit{``I’m sorry, but I couldn’t find any information about the gene associated with SNP \texttt{\{snp\}]}. Would you like to know more about SNPs in general?''} for all questions, which we simply 0.
To evaluate GeneGPT's results, we score 1 if (a) the predicted answer exactly matches the ground-truth or (b) the gene is a non-coding, and GeneGPT's answer mentions it;
we score 0.5 if there is a partial match, and 0 otherwise.

\paragraph{Disease gene location:} 
Following the evaluation of the gene disease association task in GeneTuring, we measure the recall of ground-truth chromosome locations.
We manually evaluate all answers given by the New Bing, and consider partial match as correct.
For example, if the new Bing answers ``17q21'' and the ground-truth answer is ``17q21.2'', we still consider the prediction by new Bing correct.
We automatically evaluate GeneGPT's prediction under a more strict setting where we only consider exact matches of GeneGPT as correct.

\paragraph{Sequence gene alias:}
We manually evaluate all answers predicted by GeneGPT and the New Bing, and measure the recall of the ground-truth gene aliases.
We only consider exact matches between the predicted alias and a ground truth alias, but ignore the case difference (e.g., 'Myc' and 'MYC' are still considered as a match). 

\section{Error types}
\label{appendix:error}

\paragraph{Error type 1 (E1):} Errors caused by using the wrong API or not using APIs. This only happens in disease-related tasks where the model uses the \texttt{gene} instead of the \texttt{omin} database.
One example is shown in Figure~\ref{fig:e_1}.

\begin{figure}[!ht]
    \framebox{
    \parbox{0.45\textwidth}{
    \small
    \colorbox{black!20}{\textbf{Error type 1 (E1)}} \newline \newline
    \colorbox{blue!20}{\textbf{GeneTuring question (Gene disease association)}} \newline
    What are genes related to Immunodeficiency due to defect in MAPBP-interacting protein? \newline 
    
    \colorbox{red!20}{\textbf{Wrong API call}} \newline
    [\url{https://eutils.ncbi.nlm.nih.gov/entrez/eutils/esearch.fcgi?db=gene&retmax=5&retmode=json&sort=relevance&term=Immunodeficiency\%20due\%20to\%20defect\%20in\%20MAPBP-interacting\%20protein}] \newline

    \colorbox{green!20}{\textbf{Right API call}} \newline
    [\url{https://eutils.ncbi.nlm.nih.gov/entrez/eutils/esearch.fcgi?db=omim&retmax=5&retmode=json&sort=relevance&term=Immunodeficiency\%20due\%20to\%20defect\%20in\%20MAPBP-interacting\%20protein}]
    }
    }
    \caption{An example of Error type 1 (E1). The wrong API call uses \texttt{gene} database instead of the \texttt{omin} database used by the right API call.} 
    \label{fig:e_1}
\end{figure}

\paragraph{Error type 2 (E2):} Errors caused by using the right API but wrong arguments, e.g., passing terms to \texttt{id}. One example is shown in Figure~\ref{fig:e_2}.

\begin{figure}[!ht]
    \framebox{
    \parbox{0.45\textwidth}{
    \small
    \colorbox{black!20}{\textbf{Error type 2 (E2)}} \newline \newline
    \colorbox{blue!20}{\textbf{GeneTuring question (SNP location)}} \newline
    Which chromosome does SNP rs397784008 locate on human genome? \newline 
    
    \colorbox{red!20}{\textbf{Wrong API call}} \newline
    [\url{https://eutils.ncbi.nlm.nih.gov/entrez/eutils/esummary.fcgi?db=snp&retmax=5&retmode=json&sort=relevance&id=rs397784008}] \newline

    \colorbox{green!20}{\textbf{Right API call}} \newline
    [\url{https://eutils.ncbi.nlm.nih.gov/entrez/eutils/esummary.fcgi?db=snp&retmax=5&retmode=json&sort=relevance&id=397784008}]
    }
    }
    \caption{An example of Error type 2 (E2). The wrong API call passes all \texttt{rs397784008} to the \texttt{id} argument, while the right API should only pass \texttt{397784008} without \texttt{rs} to the \texttt{id} argument.}
    \label{fig:e_2}
\end{figure}

\paragraph{Error type 3 (E3):} Errors caused by not extracting the answer in the API result, most commonly seen in gene function extraction. One example is shown in Figure~\ref{fig:e_3}.

\begin{figure}[!ht]
    \framebox{
    \parbox{0.45\textwidth}{
    \small
    \colorbox{black!20}{\textbf{Error type 3 (E3)}} \newline \newline
    \colorbox{blue!20}{\textbf{GeneHop question (SNP gene function)}} \newline
    What is the function of the gene associated with SNP rs1481036795? Let's decompose the question to sub-questions and solve them step by step. \newline 
    
    \colorbox{red!20}{\textbf{Wrong answer (scored 0.5)}} \newline
    SEPTIN11 is a filament-forming cytoskeletal GTPase. \newline \newline
    \colorbox{green!20}{\textbf{Right answer}} \newline
    SEPT11 belongs to the conserved septin family of filament-forming cytoskeletal GTPases that are involved in a variety of cellular functions including cytokinesis and vesicle trafficking (Hanai et al., 2004 [PubMed 15196925]; Nagata et al., 2004 [PubMed 15485874]).
    }
    }
    \caption{An example of Error type 3 (E3). The wrong answer only contains the protein family information without other details about its functions.} 
    \label{fig:e_3}
\end{figure}

\paragraph{Error type 4 (E4):} The model makes the right API call, but the results do not contain the answer. These questions are not answerable with the Web APIs. One example is shown in Figure~\ref{fig:e_3}.

\begin{figure}[!ht]
    \framebox{
    \parbox{0.45\textwidth}{
    \small
    \colorbox{black!20}{\textbf{Error type 4 (E4)}} \newline \newline
    \colorbox{blue!20}{\textbf{GeneTuring Question (Gene location)}} \newline
    Which chromosome is AC093802.1 gene located on human genome? \newline 
    
    \colorbox{red!20}{\textbf{Right API call returns no results}} \newline
    [\url{https://eutils.ncbi.nlm.nih.gov/entrez/eutils/esearch.fcgi?db=gene&retmax=5&retmode=json&sort=relevance&term=AC093802.1}]
    }
    }
    \caption{An example of Error type 4 (E4). The model makes the right API call but there is no entry returned from the API. The gene AC093802.1 is not indexed by the NCBI gene database.} 
    \label{fig:e_4}
\end{figure}

\paragraph{Other errors (O):} Errors that cannot be classified into E1-4 are included in this category, which are most commonly seen in BLAST-related tasks where the chromosomes are right but the specific ranges are not matched (e.g., chr8:7081648-7081782 v.s. chr8:1207812-1207946) because the original GeneTuring benchmark does not specify a reference genome.

\end{document}